\documentclass[preprint,12pt,authoryear]{elsarticle}
\usepackage{hyperref}

\usepackage{amssymb}
\usepackage{amsmath}
\usepackage{booktabs}
\usepackage{float}

\usepackage{algorithm}
\usepackage{algpseudocode}
\usepackage{comment}
\journal{Industrial Information Integration}

\begin{document}

\begin{frontmatter}



\title{ASTRO: Adaptive Spatio-Temporal Reinforcement Optimization for GNN Powered Anomly Detection in Cyber Physical Systems} 



\author[1]{Rai Ali Yar}
\ead{f223361@cfd.nu.edu.pk}

\author[2]{Umaisa Lail}
\ead{umaisalail@gmail.com}

\author[1,*]{Anwar Shah}
\ead{anwar.shah@nu.edu.pk}

\affiliation[1]{organization={Department of Computer Science, FAST NUCES},
            addressline={Development Sector}, 
            city={Faisalabad},
            postcode={}, 
            state={},
            country={Pakistan}}

\affiliation[2]{organization={Department of Information Technology, Riphah International University},
            addressline={}, 
            city={Faisalabad},
            postcode={}, 
            state={},
            country={Pakistan}}

\cortext[cor1]{Corresponding author}

\begin{abstract}
Anomaly detection in Industrial Internet of Things (IIoT) environments is essential to protect the Industrial Control Systems (ICS) and Cyber-Physical Systems (CPS) from occuring run time false data injection and other malicious attacks. The increasing complexity of sensor networks and interconnected control loops makes it difficult to identify anomalous behavior hidden within high-dimensional and time-dependent signals. To address these challenges, this article introduces Adaptive Spatio-Temporal Reinforcement Optimization ASTRO (ASTRO), a novel anomaly detection framework that pioneers the use of reinforcement learning for dynamic threshold optimization. By integrating a Deep Q-Network (DQN) with Graph Neural Networks (GNNs), temporal modelling and a Multi-Head Attention mechanism, ASTRO continuously adapts its decision boundaries to improve detection accuracy. The GNN component models the spatial relations among sensors, Temporal model captures time series dependencies and the attention layer highlights most informative time steps. The model generates continuous anomaly scores, which are transformed into binary decisions using an adaptive threshold, optimized via a Deep Q-Network (DQN). The ASTRO approach is evaluated on two real world industrial benchmarks: the Secure Water Treatment (SWaT) and Water Distribution (WADI) datasets. The proposed model achieves an exceptional performance on the SWaT with F1 score of 0.990. Moreover, on highly complex 127 end devices WADI dataset, it secures F1 score of 0.788, outperforming state-of-the-art baselines by nearly 14\%. Results across multiple runs confirm consistent generalization and stability. These experiments demonstrate that the ASTRO framework is  highly practical and scalable method for strengthening the large scale cyber physical infrastructures.
\end{abstract}




\end{frontmatter}



\section{Introduction}
The IIoT is spreading across
various sectors, such as water treatment facilities, oil and gas refining,
chemical production, and energy generation. There has been growing need for remote
business presentations, virtual assistance, and telecommuting with technological advancements.
Nowadays IIoT is getting connected to public networks more often. This helps in remote operations, but it has also increase the risk of cyber attacks. Because these attacks can cause huge money losses and also connected to human lifes, finding the anomalies inside IIoT is really important. The real time sensor data give us a continuous look at the system. But it is not just for simple status update, it actually show the complex patterns of how system behave over time. So, catching anomalies in these time series data is super necessary for IIoT safety. This make time series anomaly detection a very high priority for the research area.

  Anomaly detection algorithms are changing to meet new problems. Tradionational approaches like Angle Based Methods \cite{Anglebasedoutlier}, predictive modeling, PCA \cite{PCA}, and density estimation \cite{OutlierHighDeminesion} were pretty good at finding basic outliers. But these method often fail when we use them on complex systems where data jump around too much. Modern CPS have very tricky time patterns that classic math just can't map out well. This make it really hard to find exact anomalies. So now there is big shift going towards Deep Learning approaches \cite{impAI}\cite{surveycomprehensive} to handle such complex dynamics.

   Autoencoders (AE) use an encoder decoder setup to squeeze the data down into smaller form and then build it back again. Anomalies are found by checking reconstruction error, which is the difference between the real input and the built one \cite{AE1}\cite{AEtimeseries}\cite{AEdeepgausiananomly}\cite{AEnonlinear}. Variational Autoencoders (VAE) take this idea a step further by learning probability based on the hidden shapes. Therefore, this help them do much better in the anomaly detection in time series data by looking at the reconstruction likelihood \cite{VAEUnsupervised}\cite{VAEreconprob}. But even with these upgrades, these method often struggle to catch really tricky moving parts inside temporal data. To fix such gap, Long Short Term Memory (LSTM) models are being used instead. They give much better results, when it comes to recognize such complex dynamic behaviors \cite{lstm-AE}\cite{LSTMdync}\cite{lstmsurvey}.

   Even though deep learning make anomaly detection much better, but often struggle to map out tricky cross-dimensional relationships. To fix this problem, people start using GNNs \cite{GNNsurvey}. For example, Zekai and Dingshuo built a framework that mix graph structure learning with the graph convolutions, and they use transformer based setup, to catch how things changes over the time \cite{GNNlearning}. Inside of these frameworks, Graph Convolutional Networks (GCNs) are really needed to gather, the information across different networks. This help us get a much deeper look into how different dimensions can be connected with each other \cite{GCNAtrributednet}\cite{IOTpaper}.

  Having such capable deep learning algorithms, still gives bad F1 scores, because they make too much false positives. In sensitive area like the CPS, these mistake can be very dangerous. For example, the Stuxnet attack on the Iran nuclear plants, show us how failing to catch anomalies on time can cause a huge disaster \cite{stuxnet}. In order to get really high F1 score, we need a good classifier that can clearly tell difference between normal data and attack. So to fix this, we propose adding a DQN \cite{DQNtheory}. By using a reward system to pick the best threshold, the DQN find the maximum possible F1 score using the outputs from the Neural Network (NN) \cite{DQNdemo}.

   This study introduces a ASTRO framework. We have used, DQN based threshold optimization for anomaly detection in CPS. The whole setup mix spatial, temporal, and context learning together to understand the messy connections inside end devices data. A GCN pull out spatial correlations based on how sensors connect and Bidirectional LSTM (BiLSTM) catchs the moving time dynamics. To focus on important time feature, we use multi-head attention mechanism. Then the mixed outputs go through fully connected layers to give an anomaly score between 0 and 1. To make the classification better,then DQN agent adjust the decision threshold on the go. It learn using a reward system based on the F1-score. The DQN keep changing the threshold, it chooses to increase, decrease, or keep it same so it find the perfect balance between precision and recall.The contribution of this study is listed below:

\begin{itemize}
  \item We propose \textbf{ASTRO}, a novel  deep learning framework that jointly models spatial, temporal, and contextual dependencies to accurately detect anomalies in cyber-physical systems.
  \item We introduce a DQN optimized thresholding mechanism that adaptively selects the decision boundary using an F1-score reward formulation, ensuring optimal precision and recall under varying operational conditions.
  \item We validate the ASTRO approach empirically on the SWaT and WADI benchmark datasets, demonstrating a consistent improvement in detection capabilities and achieving an F1-score of 0.990, superseding state of the art baselines.
\end{itemize}

In the next section, related work has been discussed. Section III discuss the formulated problem. Section IV presented the proposed framework. Section V reports experimental setup and results. Finally, Section VI concludes paper and outlines the directions for future research.

\section{Related Work}

\begin{figure*}[t]
    \centering
    \includegraphics[width=0.85\textwidth]{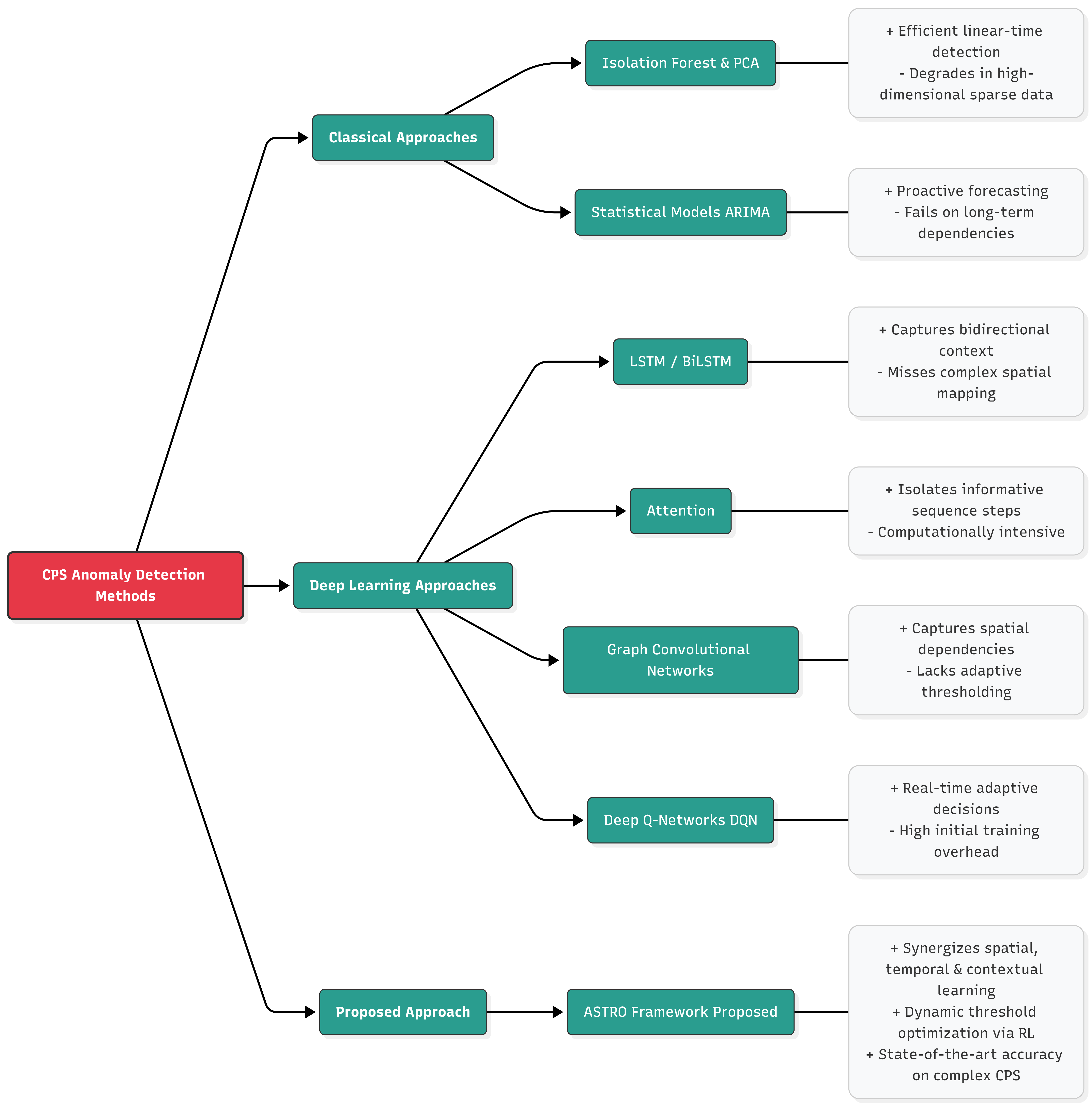}
    \caption{Taxonomy of existing Cyber-Physical Systems (CPS) anomaly detection methodologies and the positioning of the proposed ASTRO framework.}
    \label{fig:taxonomy_related_work}
\end{figure*}

Finding anomalies in modern CPS is a very big challenge because they are built so complex. Lot of research work is done in this area \cite{GNNsurvey}\cite{lstm-AE}\cite{DQNtheory}, but there are still some big research gaps left. The research papers, we have right now can mostly be splited into two main parts : the old traditional methods and the deep learning approaches. You can see this whole breakdown in Figure~\ref{fig:taxonomy_related_work}.

In the first research stream,  classic methods show strong performance in certain cases.Starting with, the Isolation Forest algorithm give a fast linear time detection way. It works by  directly isolating the anomalies, instead of learning normal behavior first. This makes it really good for the high dimensional and large scale datasets \cite{iforest}. Another common method used is the Principal Component Analysis (PCA). It finds irregularities by keep watching the projection of input data onto the orthogonal residual subspace \cite{PCA}. But, the old proximity based outlier detections, often struggle a lot in the high dimensional space. Because the data is so spread out there, moreover data sparsity makes every single point look like an outlier \cite{OutlierHighDeminesion}.

To give a proactive defense, researchers used statistical models like ARIMA to guess network behaviors and give early warning for such possible attacks \cite{ARIMA}. Also, frameworks like MERLIN \cite{MERLIN} and SAND \cite{SAND} try to automate time series anomaly detection. They can do this without needing too much parameter tuning by using the statistical modeling and sub dataset weighting. The performance of these classic methods, starts to drop, when the time series get more dimensions. This can lead to the lower accuracy. These ways gave us a good starting point, but in the end they just do not have the understanding of the long term dependencies or keep the contextual relationships between many different time series data streams.

Over the last decade, research has moved a lot toward deep learning ways, because they have always performed better than the old classic techniques. These models use neural network setup to catch long term time patterns in data much better. For example, the LSTM design show a lot of success in learning both short and long term connections. This helps us to find anomalies across different time gaps \cite{LSTMdync}\cite{lstm-AE}. Building on this, BiLSTM algorithm \cite{BiLSTM} make normal LSTM even better by reading the sequences forward and backward. This way, the model can catch context information from both the past and future time steps at the same time. This really improves the detection performance in sequenced data stream \cite{IIOTlstm}.

At the same time, the Transformer architecture \cite{Transformer} changed sequence modeling completely by bringing in the great self-attention mechanisms. It let's the model catch global dependencies, without being stuck by the recurrent or convolutional structures. This new idea led people to add attention mechanisms into the time series models, so they can pick out the most informative time steps inside a sequence. Mixing the attention with BiLSTM makes the model ability to focus on critical temporal features much better. Also, the multi-head attention let the system look at information from many different representation of the subspace at once. This helps to catch a wide mix of temporal dependencies. Recent tests showed that, the Temporal Fusion Transformers (TFT) \cite{TFT}, Informer \cite{Informer} and the Anomaly Transformer \cite{AnomalyTransformer}, have demonstrated, that multi-head attention is effective at finding the long term dependencies and capturing complex temporal patterns, ultimately improving accuracy in time series forecasting and anomaly detection. They have given very high performance on the large scale and complex time datasets.

Even with all these new upgrades, the challenge to map out complex structured data was still there. So this brings us in the GNNs. GCNs, do the convolution math directly on the graph structured data. They are really good at catching both single node features and local shape information to get best results in different semi-supervised task \cite{GCN}. Adding to this, the Spatio-Temporal Graph Convolutional Networks (STGCN) \cite{STGCN} mixes graph convolutions with the time modeling. This help them to handle, both the spatial connection between nodes and the moving parts of sequence data at the very same time. Just recently, the GNN based anomaly detection frameworks \cite{GDN} start using these graph shape to model, the tricky sensor to sensor relationship inside Cyber-Physical Systems. This makes it very easy to find both context and the collective anomalies.

Recent research work in the CPS anomaly detection have moved beyond static thresholding and old supervised classification. Now it is shifting toward adaptive decision making techniques, that can adjust operating threshold in real time \cite{BDQN}. Latest studies have added in DQN based framework\cite{DQN}. They treat the open set recognition problem as a Markov Decision Process (MDP). This way help get a more detailed and adaptive classification of known network traffic. At the same time, it really improve how we catch new unseen cyber attacks.

Looking at all the current literature show us that graph based learning, time modeling, and adaptive decision making have all grow on their own. But putting them all together into one single framework is still unexplored. But, the good thing about our proposed approach is not just that it is a hybrid mix, it is actually about its strong technical design and validated performance. As far as we know, this ASTRO framework is the very first one to mix GCNs with temporal modelling, so it can catch spatial and temporal dependencies at the same time. And it also add a DQN for dynamic threshold optimization. This teamwork let the model beat static methods. Our novelty lies in a graph construction approach that models both inter-machine and intra-machine relationships. It give a much more accurate, adaptive, and easy to understand solution for finding anomalies inside complex CPS environments.

\section{Problem Formulation}
Let say a CPS is represented by a spatial graph $\mathcal{G} = (\mathcal{V}, \mathcal{E})$. Here $\mathcal{V}$ is the set of $N$ sensor and actuator node, and $\mathcal{E}$ show the physical topology. For a given time window $T$, the system make multivariate sequence data. We denote this as $\mathbf{X} \in \mathbb{R}^{N \times F \times T}$, where $F$ is the number of feature per node.

We formulate the anomaly detection problem as a mapping from the spatio-temporal input space to a continuous anomaly score $\hat{y} \in [0, 1]$. We want to learn an optimal mapping function $f_{\theta}$:

\begin{equation}
    \hat{y} = f_{\theta}(\mathbf{X}, \mathcal{G})
\end{equation}

where $\theta$ represent the learnable parameters of the system.

Also, to classify a temporal sequence as normal ($c=0$) or anomalous ($c=1$), we need to apply a decision threshold $\tau \in [0, 1]$ to the continuous score $\hat{y}$:

\begin{equation}
    c = 
    \begin{cases} 
      1, & \text{if } \hat{y} > \tau \\ 
      0, & \text{otherwise} 
    \end{cases}
\end{equation}

Because static threshold fail to adapt to different attack distribution over time, our second goal is to treat threshold selection as a dynamic optimization problem. The goal is to find an optimal threshold policy $\tau^{*}$ that maximize a predefined performance reward $R$ (like the $F_{1}$-score) over a validation set:

\begin{equation}
    \tau^{*} = \arg\max_{\tau} R(\tau, \hat{y}, y_{true})
\end{equation}

where $y_{true}$ show the ground truth labels. So, the big goal of this research is to jointly optimize the mapping parameters $\theta$ for accurate spatio-temporal feature extraction and the decision boundary $\tau$ for robust binary classification at the same time.

\begin{figure*}[htbp]
    \centering
    \includegraphics[width=\textwidth]{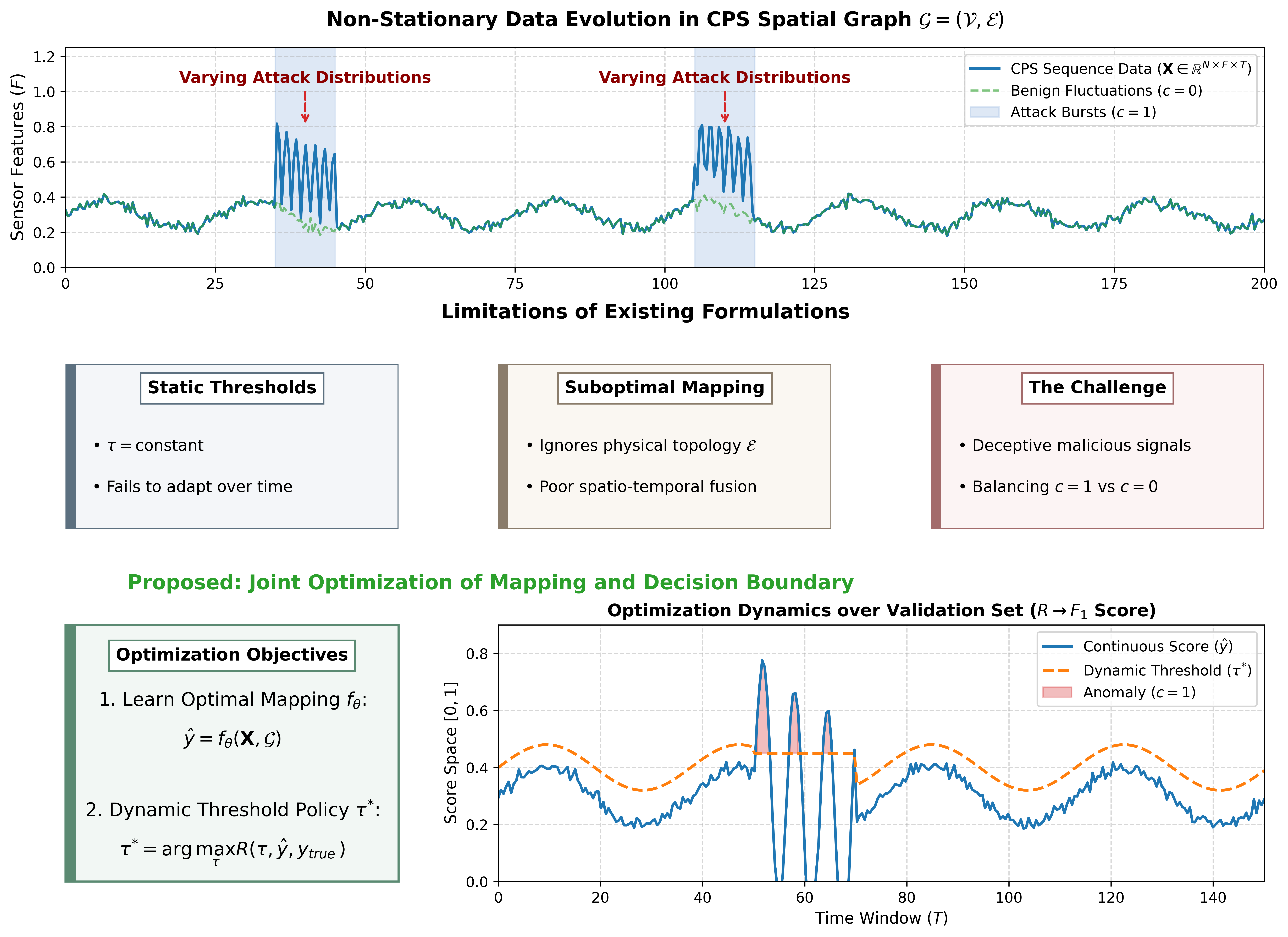}
    \caption{Visual summary of the problem formulation. Top: Non-stationary CPS data evolution exhibiting varying attack distributions. Middle: Limitations of traditional static and isolated anomaly detection methods. Bottom: The proposed joint optimization of the spatio-temporal mapping ($f_{\theta}$) and the dynamic decision boundary ($\tau^{*}$).}
    \label{fig:problem_formulation}
\end{figure*}

As you can see in Figure \ref{fig:problem_formulation}, this joint optimization directly fix the big challenges of non-stationary industrial environments. The top panel show how stealthy attacks hide as varying distributions inside normal benign signal fluctuations. The middle panel point out the limits of existing frameworks. It show how rigid static thresholds ($\tau = \text{constant}$) are, and also show the suboptimal mapping that just ignore the physical topology ($\mathcal{E}$). Finally, the bottom panel show our proposed dynamic threshold policy ($\tau^{*}$). This policy adapt to the continuous anomaly score ($\hat{y}$) in real time so it can stabilize the decision boundary and get the maximum classification reward.

\section{Adaptive Spatio-Temporal Reinforcement Optimization (ASTRO)}
\begin{figure*}
    \centering
    \includegraphics[width=0.95\linewidth]{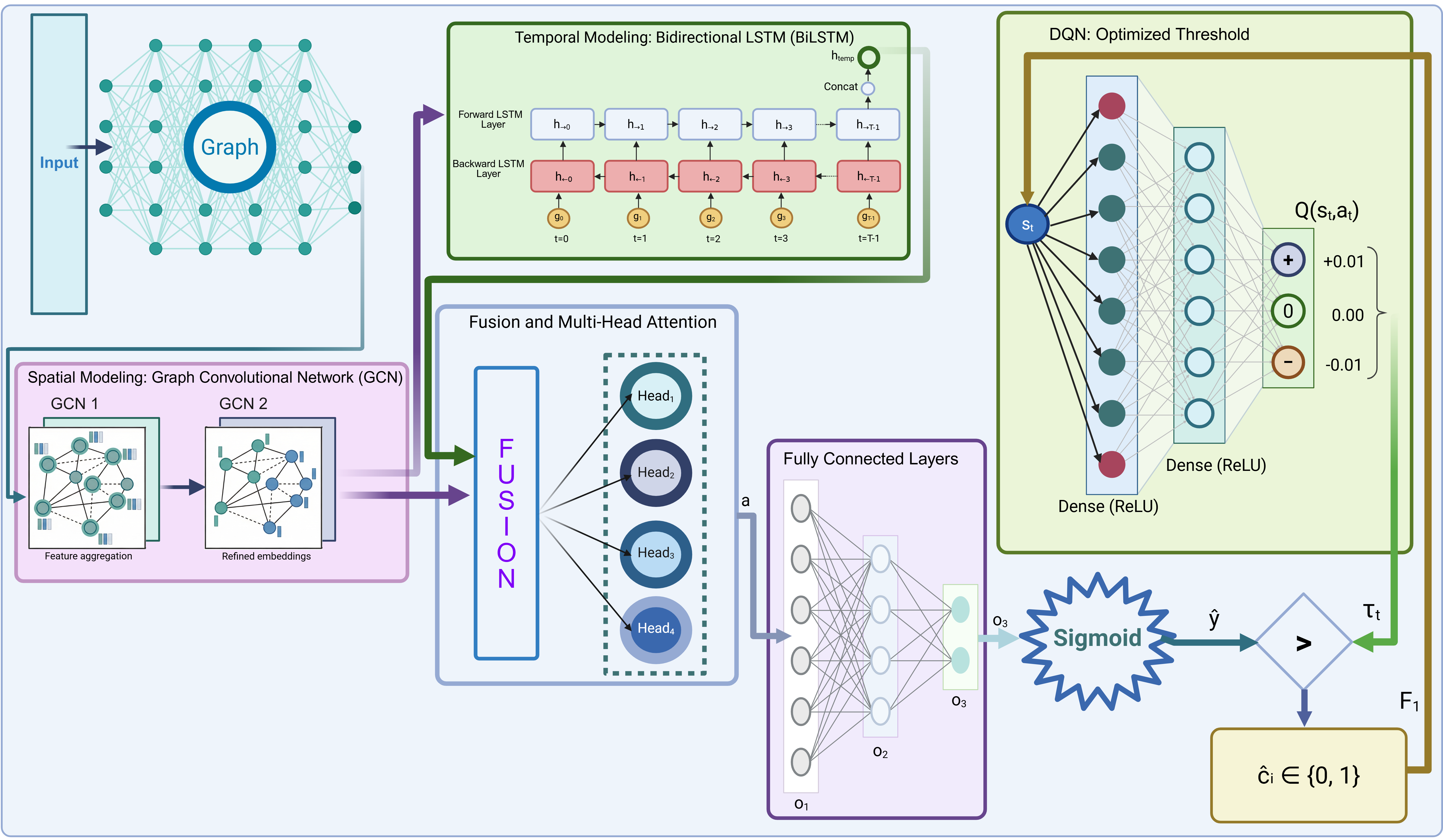}
    \caption{The proposed Framework: ASTRO }
    \label{fig:framework}
\end{figure*}
The proposed framework, named ASTRO, detects anomalies in cyber-physical systems by jointly modeling spatial and temporal dependencies in  sensors, actuators and other end devices data. As illustrated in Figure~\ref{fig:framework} and detailed in Algorithm~\ref{alg:algo1} , the architecture integrates a GCN, a BiLSTM  and a Multi-Head Attention mechanism, followed by a DQN based threshold optimization.The GCN captures spatial correlations among the end devices through graph structured aggregation, while the temporal models the time series patterns over time. Attention module further refines the temporal representation by emphasizing critical time steps that indicate abnormal behavior.Finally, a fusion layer produces an anomaly score for each sequence and the DQN adaptively tunes decision threshold to maximize detection performance across varying operating conditions.

\subsection{Graph Construction and Spatial Modeling}
\begin{figure*}[h]
    \centering
    \includegraphics[width=1\linewidth]{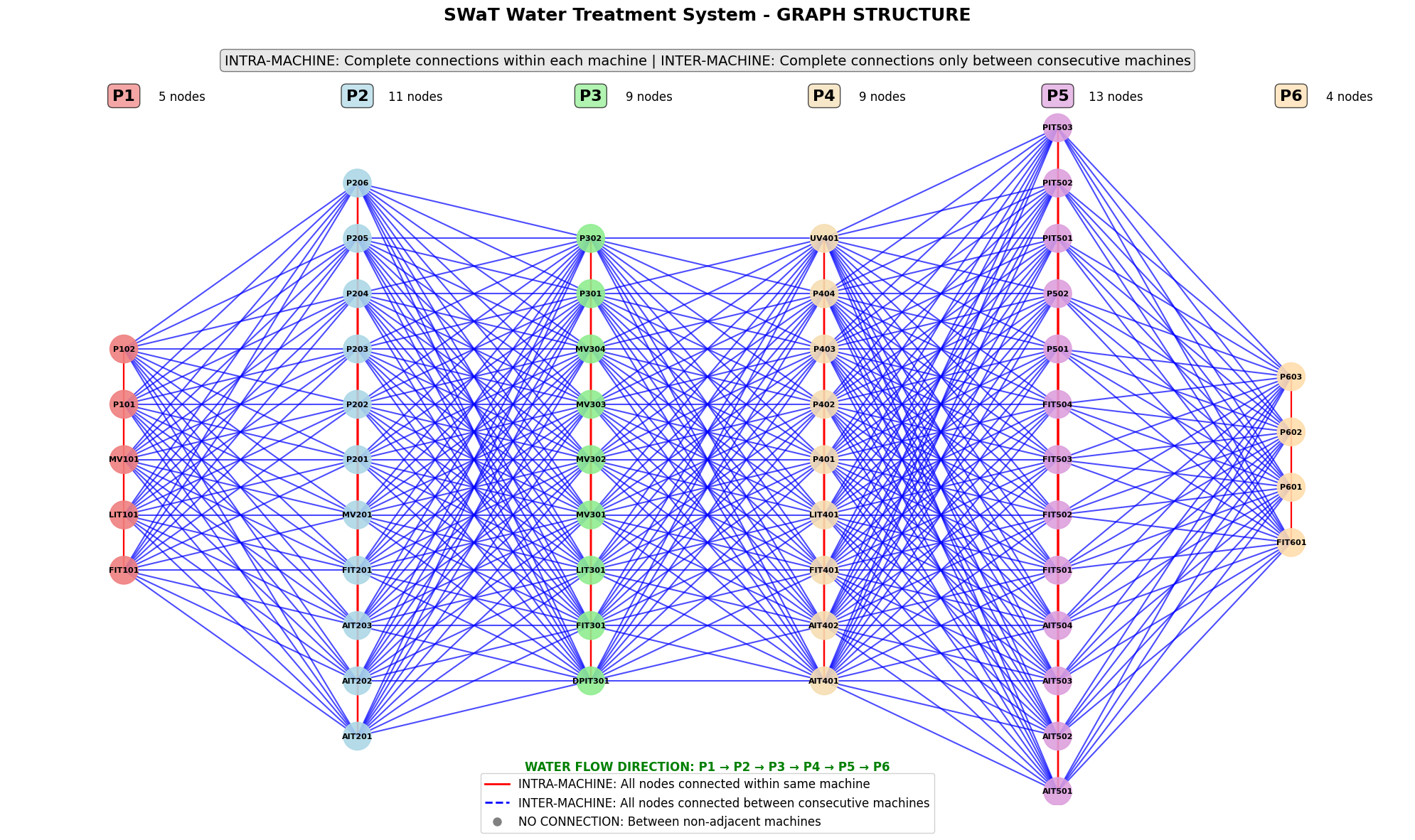}
    \caption{Swat Dataset relationships mapped  inter-machine and intra-machine on this graph model is trained}
    \label{fig:graphmodel}
\end{figure*}

As illustrated in Fig.~\ref{fig:graphmodel}, the system is represented as a undirected graph, where nodes correspond to sensors and actuators, and edges encode physical relationships. Two connection types are defined: intra-machine links (fully  connected within the same subsystem) and inter-machine links ( fully connected with the adjacent subsystems). Edge attributes encode connection strength. The graph topology is stored in tensor form as an edge index and edge attribute vector:
\[
\text{edge\_index}\in\mathbb{Z}^{2\times E},\qquad \text{edge\_attr}\in\mathbb{R}^{E\times 1},
\]
where $E$ denotes the number of directed edges

Input data are organized as sliding windows and fed to the model with shape
\[
X\in\mathbb{R}^{B\times N\times F\times T},
\]
where $B$ is the batch size, $N$ is the number of nodes, $F$ is the number of features per node (including engineered temporal features), and $T$ is the window length. For each time step $t\in\{0,\dots,T-1\}$ the node features are first projected by a small MLP to a fixed per-node embedding:
\[
H_t^{\text{MLP}} = \mathrm{MLP}(X_{:, :, :, t})\in\mathbb{R}^{B\times N\times D_{\text{in}}},
\]
with $D_{\text{in}}=32$ in the implementation.

Following the MLP, the per-batch node tensor is flattened and passed to the GCN layers exactly as implemented:
\[
H_t^{\text{flat}} = \mathrm{reshape}\big(H_t^{\text{MLP}},(B\cdot N,\,D_{\text{in}})\big),
\]
\[
G_t^{(1)} = \mathrm{ReLU}\big(\mathrm{GCNConv}_1(H_t^{\text{flat}},\;\text{edge\_index},\;\text{edge\_attr})\big),
\]
\[
G_t^{(2)} = \mathrm{ReLU}\big(\mathrm{GCNConv}_2(G_t^{(1)},\;\text{edge\_index},\;\text{edge\_attr})\big),
\]
and the GCN output is reshaped back to per-graph node embeddings:
\[
G_t = \mathrm{reshape}\big(G_t^{(2)},(B,\,N,\,D_g)\big).
\]
Each time-step graph is summarized by mean pooling across nodes to obtain a graph-level vector:
\[
\mathbf{g}_t = \frac{1}{N}\sum_{i=1}^{N} G_t[:, i, :]\in\mathbb{R}^{B\times D_g}.
\]
Stacking these pooled vectors across the window produces the temporal input for the sequence model:
\[
G_{\text{seq}} = [\mathbf{g}_0, \mathbf{g}_1, \dots, \mathbf{g}_{T-1}]\in\mathbb{R}^{B\times T\times D_g}.
\]

This description matches the code behavior: the same \texttt{edge\_index} / \texttt{edge\_attr} tensors (defined for an $N$-node graph) are reused for each time step and for the flattened batch passed into \texttt{GCNConv}. The resulting graph-level sequence $G_{\text{seq}}$ is then processed by the temporal module described in the next subsection.

\subsection{Temporal Fusion and Attention Module}
The temporal evolution of the graph-level representations is modeled using a bidirectional LSTM followed by a lightweight attention-based fusion stage. The sequence of pooled graph embeddings obtained from the previous stage is as following,
\[
G_{\text{seq}} \in \mathbb{R}^{B\times T\times D_g},
\]
is processed by a single-layer bidirectional LSTM:
\[
H_{\text{LSTM}} = \mathrm{Temporal}(G_{\text{seq}}),
\]
where the hidden size is $D_h=32$ per direction, the output corresponding to the final time step, will be concatenated from both directions of the model, which will form the temporal summary:
\[
\mathbf{h}_{\text{temp}} = H_{\text{LSTM}}[:, -1, :] \in \mathbb{R}^{B\times 2D_h}.
\]

In parallel, the last graph embedding of the sequence $\mathbf{g}_{T-1}$ is retained as the final spatial summary. These two vectors are then concatenated to yield unified representation:
\[
\mathbf{z}_{\text{fuse}} = [\,\mathbf{g}_{T-1} \,\Vert\, \mathbf{h}_{\text{temp}}\,] \in \mathbb{R}^{B\times(D_g+2D_h)}.
\]
Linear projection will reduce the dimensionality to 64:
\[
\mathbf{z}' = \mathrm{ReLU}(W_f \mathbf{z}_{\text{fuse}} + b_f).
\]

To enhance the discriminative features weighting, a multi-head attention block with four heads is applied as following:
\[
\mathbf{a} = \mathrm{MHA}(\mathbf{z}', \mathbf{z}', \mathbf{z}'),
\]
Where the attention mechanism will dynamically scale feature contributions across fused components. The attended vector $\mathbf{a}$ is then propagated through three fully connected layers:
\begin{align}
\mathbf{o}_1 &= \mathrm{ReLU}(W_1 \mathbf{a} + b_1), \\
\mathbf{o}_2 &= \mathrm{ReLU}(W_2 \mathbf{o}_1 + b_2), \\
\mathbf{o}_3 &= \mathrm{ReLU}(W_3 \mathbf{o}_2 + b_3),
\end{align}

followed by a final sigmoid classifier:
\[
\hat{y} = \sigma(W_{\text{final}}\mathbf{o}_3 + b_{\text{final}}),
\]
where $\hat{y}\in[0,1]$ denotes the anomaly probability for input window.

This design clearly mixes both spatial and temporal dependencei together. The temporal model catch the dynamic dependencies inside these moving graph states over time. Then the attention mechanism cleans up the mixed features, so it can highlight the important temporal-spatial patterns. This make it easy to tell the difference between a real attack sequence and normal behavior. Overall, this setup led to a very stable convergence for the model, and it give really strong anomaly discrimination across our whole test set.
\subsection{DQN-Based Threshold Optimization}
Instead of relying on one fixed anomaly threshold, which can underperform under varying data distributions, this paper work employs a DQN agent to adaptively optimize the decision boundary based on the model outputs. The anomaly probabilities $\hat{y}$ produced by the model serve as the input for threshold tuning.

The threshold selection task is formulated as a discrete-time MDP. At each episode $t$, the environment state $s_t$ corresponds to the current threshold $\tau_t \in [0,1]$, and the agent selects one of three possible actions:
\[
a_t \in \{-0.01, 0, +0.01\},
\]
representing a decrease, no change, or increase in the threshold, respectively. The new threshold is updated as:
\[
\tau_{t+1} = \mathrm{clip}(\tau_t + a_t, 0, 1).
\]

After each action, the anomaly labels are re-evaluated as:
\[
\hat{y}_i = 
\begin{cases}
1, & \text{if } \hat{p}_i > \tau_t,\\
0, & \text{otherwise},
\end{cases}
\]
and the agent receives a reward $r_t$ defined as the current $F_1$-score computed between the predicted labels $\hat{y}$ and the ground truth $y$. The DQN agent learns to maximize the expected cumulative reward by updating its Q-value function:
\[
Q(s_t, a_t) \leftarrow (1-\alpha)Q(s_t, a_t) + \alpha \big(r_t + \gamma \max_{a'} Q(s_{t+1}, a')\big),
\]
where $\alpha$ is the learning rate and $\gamma$ is the discount factor. The network is implemented as a three-layer fully connected neural model with ReLU activations and trained using mean squared error loss.

Through this iterative exploration of threshold, the agent converges to optimal threshold $\tau^*$ that yields the highest $F_1$-score on the validation data:
\[
\tau^* = \arg\max_{\tau} F_1(\tau).
\]
This adaptive optimization enables the anomaly detection system to dynamically adjust sensitivity according to the anomalies and maintaining stable performance even under distribution shifts or varying attack intensities. In our experiments, the DQN-optimized threshold consistently improved both precision and recall compared to static thresholding approaches.

\subsection{Algorithm }
The complete execution pipeline of the proposed ASTRO framework, integrating the spatial-temporal learning and the reinforcement learning-based threshold tuning, is outlined in Algorithm~\ref{alg:algo1}. The input to the algorithm includes the multi-dimensional sliding windows of sensor data $X$, the graph topological structures represented by the edge indices and attributes, and the designated hyperparameters for both the deep learning model and the DQN agent. The final outputs of the algorithm are the trained ASTRO model, the optimized decision boundary $\tau^*$, and the binary anomaly predictions $\hat{y}$.

\begin{algorithm}[H]
\caption{ ASTRO: Training}
\label{alg:algo1}
\renewcommand{\algorithmicrequire}{\textbf{Input:}}
\renewcommand{\algorithmicensure}{\textbf{Output:}}
\begin{algorithmic}[1] 
\Require \textbf{Data:} $X \in \mathbb{R}^{M \times N \times F \times T}$ \\
\textbf{Graph:} $edge\_index$, $edge\_attr$ \\
\textbf{Hyperparameters:} $B, D_{in}, D_g, D_h, H, \alpha, \gamma, \tau_0, E_{\text{model}}, E_{\text{dqn}}$
\Ensure Labels $\hat{y} \in \{0,1\}^{M}$ (Trained model), $\tau^{*}$ (Optimized threshold) 

\State \textbf{Preprocess:} standardize features; build sliding windows
\State Initialize model and DQN parameters

\For{epoch $= 1$ \textbf{to} $E_{\text{model}}$}
    \For{\textbf{each} training batch}
        \For{$t = 0$ \textbf{to} $T-1$}
            \State Per-node projection via MLP
            \State Apply GCN layers $\rightarrow$ per-node embeddings
            \State Pool node embeddings $\rightarrow g_t$
        \EndFor
        \State $G \leftarrow [g_0,\dots,g_{T-1}]$; feed $G$ to temporal model
        \State Fuse temporal model output with $g_{T-1}$; apply multi-head attention and FC
        \State Compute probabilities $p$; update model via weighted BCE loss
    \EndFor
\EndFor

\State Initialize DQN replay buffer; set $\tau \leftarrow \tau_0$

\For{episode $= 1$ \textbf{to} $E_{\text{dqn}}$}
    \For{\textbf{each} validation step}
        \State Select action $a \in \{-\xi, 0, +\xi\}$ ($\epsilon$-greedy)
        \State $\tau' \leftarrow \text{clip}(\tau + a, 0, 1)$
        \State Convert $p$ to labels using $\tau'$
        \State Reward $r \leftarrow \text{F1}(\hat{y}, y_{\text{val}})$
        \State Store $(\tau, a, r, \tau')$; update Q-network
        \State $\tau \leftarrow \tau'$
    \EndFor
\EndFor

\State $\tau^{*} \leftarrow$ best validation threshold observed
\State \textbf{Inference:} compute $p$ on test set; apply $\tau^{*}$ to obtain $\hat{y}$
\State \Return $\hat{y}$ (Trained model), $\tau^{*}$ (Optimized threshold)
\end{algorithmic}
\end{algorithm}

The algorithm in line 1 starts by standardizing the raw multivariate data and constructing the sliding windows to capture continuous sequences. In line 2, the weights of the ASTRO framework. The algorithm then enters the primary model training phase, represented by the outer loop in lines 3 to 14. For each batch of training data, an inner loop (lines 5 to 9) iterates through every discrete time step $t$ within the window length $T$. During this step, the node features are projected via a Multi Layer Perceptron (MLP), and the GCN layers aggregate the spatial correlations among the interconnected sensors to generate per node embeddings. These embeddings are then mean pooled to form a comprehensive graph-level summary vector $g_t$ for that specific time step. 

Next, in line 10, the sequential graph embeddings are stacked to form the temporal sequence $G$, which is fed into the temporal model to model the time dependent dynamics. In line 11, the final spatial summary is fused with the temporal output, and the multi-head attention mechanism is applied to dynamically weight the most critical temporal features. The attended representation is passed through the fully connected (FC) layers to compute the continuous anomaly probabilities $p$. Line 12 computes the weighted Binary Cross Entropy (BCE) loss and updates the model weights via backpropagation.

Lines 15 to 25 detail the threshold optimization phase using the DQN agent. In line 15, the DQN replay buffer is initialized, and the decision threshold is set to a baseline $\tau_0$. The algorithm enters the reinforcement learning loop for $E_{dqn}$ episodes. During each validation step (lines 17 to 24), the agent selects an action $a$ using an $\epsilon$-greedy policy to either increase, decrease, or maintain the threshold by a step of $\Delta$. In line 19, the new threshold $\tau'$ is computed and strictly clipped within the $[0,1]$ bounds. Line 20 applies this new threshold to the model's continuous probabilities $p$ to generate hard binary labels. In line 21, the environment returns a reward $r$ based on the $F_1$-score of these predictions against the validation ground truth. The agent stores this transition experience and updates the Q-network in line 22, subsequently transitioning to the new state in line 23. 

Finally, lines 26 to 28 provide the realization of the testing and inference phase. The algorithm extracts the best performing threshold $\tau^*$ observed during the DQN validation episodes. In line 27, the trained ASTRO model computes the anomaly probabilities on the unseen test set, and the optimized threshold $\tau^*$ is applied to obtain the final binary classification labels. The algorithm concludes by returning the trained hybrid model, the optimal threshold, and the final predictions.

The overall time complexity of the ASTRO framework is primarily governed by the spatial temporal feature extraction phase and the subsequent reinforcement learning optimization. For the deep learning architecture, processing a single data sequence of length $T$ with $N$ nodes and $L$ edges requires $O(T(L \cdot D_g + N \cdot D_g^2))$ operations for the Graph Convolutional Network layers, where $D_g$ is the hidden dimension of the graph network. The temporal sequence modeling via the temporal modelling and multi head attention mechanisms adds a complexity of $O(T \cdot D_h^2 + T^2 \cdot D_h)$, where $D_h$ represents the recurrent hidden dimension. Therefore, over $E_{model}$ epochs and $M$ total training samples, the model training complexity scales as $O(E_{model} \cdot M \cdot (T(L + N \cdot D_g^2) + T \cdot D_h^2 + T^2 \cdot D_h))$. Following the feature extraction, the DQN threshold tuning operates over $E_{dqn}$ episodes and $V$ validation steps, contributing an additional lightweight complexity of $O(E_{dqn} \cdot V)$. Because the physical graph topology remains static and the sliding window length $T$ is kept small, the dominating factors scale linearly with the dataset size. This ensures that the training process remains computationally efficient and the inference time is kept minimal, making the framework highly viable for real time cyber physical system monitoring.

To make the evaluation process formal, Algorithm \ref{alg:astro_testing} show the exact inference phase of the ASTRO framework. The process start by deploying the fully trained spatial temporal model together with the optimal decision boundary $\tau^*$ that the Deep Q Network found. During inference, the model run strictly in evaluation mode and it process unseen test sequence batch by batch. For each sequence, important feature are pulled out through the integrated spatio temporal and Multi Head Attention layers. This finally give a continuous anomaly probability score between zero and one. 

In the next decision step, these continuous scores are binarized using that dynamically tuned threshold. Any sequence that give a probability equal to or higher than $\tau^*$ is marked as a malicious attack. But if the score is lower, it is confirmed as normal behavior. At the end, these binary predictions are directly checked against the ground truth labels to count exactly how many True Positives, True Negatives, False Positives, and False Negatives we got.

\begin{algorithm}[H]
\caption{ASTRO: Inference and Evaluation}
\label{alg:astro_testing}
\renewcommand{\algorithmicrequire}{\textbf{Input:}}
\renewcommand{\algorithmicensure}{\textbf{Output:}}
\begin{algorithmic}[1] 
\Require Test Data: $\mathbf{X}_{test} \in \mathbb{R}^{M_{test} \times N \times F \times T}$, True Labels $Y_{test}$
, Graph Topology: $\mathcal{G} = (\mathcal{V}, \mathcal{E})$
\Require Trained Model: $f_{\theta}$, Optimized Threshold: $\tau^{*}$
\Ensure Binary predictions $\hat{Y}$, Performance Metrics (Accuracy, Precision, Recall, F1 score)

\State Initialize empty probability list $P = [~]$
\State Set model $f_{\theta}$ to evaluation mode (disable gradient computation)

\For{\textbf{each} batch $(\mathbf{X}_{batch}, y_{batch})$ \textbf{in} $\mathbf{X}_{test}$}
    \State Extract spatial temporal features 
    \State Compute logits and apply sigmoid to obtain continuous anomaly scores: $p = \sigma(f_{\theta}(\mathbf{X}_{batch}, \mathcal{G}))$
    \State Append $p$ to $P$
\EndFor

\State Initialize empty prediction list $\hat{Y} = [~]$
\For{\textbf{each} score $p_i \in P$}
    \If{$p_i \ge \tau^{*}$}
        \State $\hat{y}_i \leftarrow 1$ \Comment{Classified as Attack / Anomaly}
    \Else
        \State $\hat{y}_i \leftarrow 0$ \Comment{Classified as Normal}
    \EndIf
    \State Append $\hat{y}_i$ to $\hat{Y}$
\EndFor

\State Compute True Positives (TP), True Negatives (TN), False Positives (FP), and False Negatives (FN) by comparing $Y_{test}$ with $\hat{Y}$
\State Calculate Precision, Recall, F1 score, and Accuracy
\State \Return $\hat{Y}$, Calculated Metrics
\end{algorithmic}
\end{algorithm}
\section{Experiments}
This section give a full evaluation of our proposed ASTRO framework to show how well it had performed at finding anomalies inside CPS. Intially, we described the industrial dataset that we used for our training and testing. Next, we outline the standard evaluation metrics we used to strictly check the model performance on different metrics. After that, we have provided details about the whole experimental setup. Finally, we give a comparative visual analysis using explainable AI (XAI) techniques, specifically SHAP, LIME, Grad-CAM, and Counterfactual explanations. This help us understand how the model make its decisions and prove its strong detection capabilities.

\subsection{Dataset and Evaluation Protocol}

To make sure we have a strong evaluation across different scale and topology, we use two real world industrial control system testbeds: the SWaT dataset and the WADI dataset.

The SWaT dataset \cite{swatdataset} is a real industrial control system testbed, and we use it to test our proposed framework. We take two official log files: a fully normal (clean) dataset and a mixed dataset that contain both normal and attack samples. The fully normal dataset have 496,800 normal and 0 attack records across 51 different sensors and actuators data. The mixed dataset have 396,681 normal and 53,283 attack records across the same 51 sensors and actuators data as for the fully normal dataset, difference is that both dataset files are recorded on two separate time spans.

To test how well the framework scale on a larger and more complex topology, we also use the WADI dataset \cite{wadidataset}. The WADI testbed cover a much larger physical network. It consist of 127 dimensions that represent different sensors and actuators. The dataset is split into a training set that hold 118,795 normal operational records and a testing set with 17,275 records. When you compare it to SWaT, WADI give a highly imbalanced and very challenging detection scenario. This is because the anomalous attack sequences make up only 5.99 percent of the testing data.

To ensure balanced training, a mixed dataset file was chosen, and an equal number of attack and normal samples were selected and divided into training (70\%), validation (15\%), and test (15\%) sets. Each sequence represented one second of operation with its corresponding multivariate readings. After training, the clean dataset was used for independent validation of false-positive behavior in unseen normal conditions.

To assess classification performance, we calculate Precision, Recall, the F1 score, and Accuracy based on True Positives (TP), True Negatives (TN), False Positives (FP), and False Negatives (FN). Due to the severe class imbalance inherent in industrial datasets, the F1 score serves as our primary benchmark metric. These metrics are defined in table~\ref{tab:metricstable}.

\begin{table}[htbp]
\centering
\renewcommand{\arraystretch}{1.5} 
\caption{Evaluation metrics formulation for the ASTRO}
\label{tab:metricstable}
\begin{tabular}{ll}
\toprule
\textbf{Metric} & \textbf{Equation} \\
\midrule
Precision & $Precision = TP / (TP + FP)$ \\
Recall    & $Recall = TP / (TP + FN)$ \\
F1 Score  & $F1 = 2 \times (Precision \times Recall) / (Precision + Recall)$ \\
Accuracy  & $Accuracy = (TP + TN) / (TP + TN + FP + FN)$ \\
\bottomrule
\end{tabular}
\end{table}
\subsection{Experimental Setup}
All experiments were conducted on a high performance laptop based workstation. Further detials of the setup and hardware components are mentioned in table~\ref{tab:exp_setup}.

\begin{table}[htbp]
\centering
\caption{Experimental setup and hardware specifications.}
\label{tab:exp_setup}
\begin{tabular}{ll}
\toprule
\textbf{Component} & \textbf{Specification} \\
\midrule
CPU                   & Intel Core i7-11850H (8 cores, 16 threads, 2.5 GHz) \\
RAM                   & 32 GB \\
GPU                   & NVIDIA RTX A3000 Laptop GPU \\
Storage               & Solid-state drive (SSD) \\
Deep Learning Library & PyTorch \\
Hardware Acceleration & CUDA (GPU-accelerated training and inference) \\
\bottomrule
\end{tabular}
\end{table}

The hyperparameter configurations used to train the ASTRO model are summarized in Table~\ref{tab:hyperparameters}. The network was trained for 30 epochs using a batch size of 64. To update the network weights, we employed the Adam optimizer with a learning rate of $1 \times 10^{-3}$. Furthermore, because anomalies constitute a minority of the data, a weighted Binary Cross-Entropy (BCE) loss function was utilized during optimization to heavily penalize misclassifications of the attack class, thereby mitigating the residual effects of class imbalance.

\begin{table}[hbt!]
\centering
\renewcommand{\arraystretch}{1.3} 
\caption{Hyperparameter Configurations for the ASTRO Model}
\label{tab:hyperparameters}
\begin{tabular}{llc}
\toprule
\textbf{Symbol} & \textbf{Parameter Description} & \textbf{Value} \\
\midrule
$N$ & Number of Graph Nodes (Sensors/Actuators) & 51 \\
$F$ & Features per Node & 1 \\
$T$ & Window Length (Sequence Time Steps) & 10 \\
$D_g$ & GCN Hidden Dimension & 32 \\
$D_h$ & BiLSTM Hidden Dimension (per direction) & 32 \\
$B$ & Batch Size & 64 \\
$\alpha$ & Learning Rate & $1 \times 10^{-3}$ \\
$E_{\text{model}}$ & Training Epochs & 30 \\
\bottomrule
\end{tabular}
\end{table}

\subsection{Performance on Mixed Test Set}
Using the DQN-optimized decision threshold ($\tau = 0.51$), the model achieved outstanding detection results on the balanced test subset. The confusion matrix shown in the Table~\ref{tab:confusion} summarizes outcomes between normal and attack data.

\begin{table}[h]
\centering
\caption{Confusion matrix on mixed test set}
\label{tab:confusion}
\begin{tabular}{lcc}
\toprule
 & \textbf{Predicted Normal} & \textbf{Predicted Attack} \\
\midrule
\textbf{Actual Normal} & 7909 & 80 \\
\textbf{Actual Attack} & 77 & 7909 \\
\bottomrule
\end{tabular}
\end{table}

The corresponding performance metrics are summarized in Table~\ref{tab:metrics}. The model reached an F1-score of 0.9902, indicating nearly perfect discrimination between normal and anomalous states.

\begin{table}[h]
\centering
\caption{Performance metrics on test data}
\label{tab:metrics}
\begin{tabular}{lc}
\toprule
\textbf{Metric} & \textbf{Value} \\
\midrule
Accuracy & 0.9902 \\
Precision & 0.9831 \\
Recall & 0.9975 \\
F1-Score & 0.9902 \\
\bottomrule
\end{tabular}
\end{table}

\subsection{Generalization to Clean Data}
\begin{figure}
    \centering
    \includegraphics[width=1\linewidth]{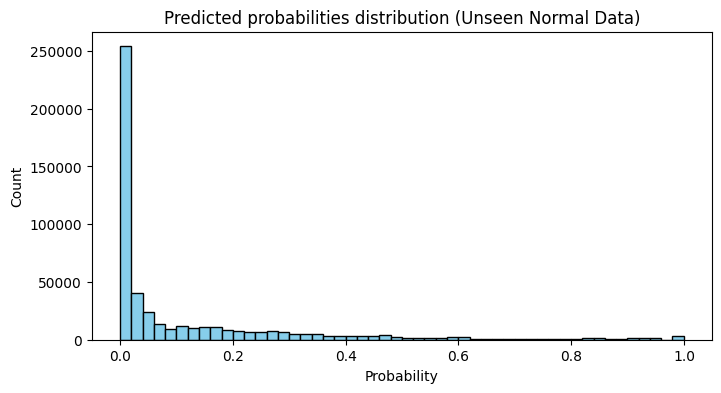}
    \caption{Predicted probabilities distribution chart on the unseen data checked on the trained model, x-axis shows the count of the records, normal records near 0 probability and attack/anomaly near   1 probability }
    \label{fig:unseendata}
\end{figure}

To evaluate robustness under realistic operating conditions, the trained model was tested on the fully normal SWaT log containing 496,800 time steps. At the learned threshold ($\tau=0.51$), approximately 20,000 false positives were observed, corresponding to less than 4\% of all samples were false  positive. Figure~\ref{fig:unseendata} illustrates the predicted probability distribution, showing that most false positives occur near the lower probability boundary, confirming that the model assigns high confidence to normal behavior in unseen scenarios. This supports the F1 score that our results give, around 96\%  true positives results on the unseen data for the trained model.

\subsection{Baseline Comparison}
To properly contextualize the effectiveness of our proposed ASTRO framework alongside the DQN threshold tuning, we compared our results with several previously reported baselines from the literature across both the SWaT and WADI datasets. These benchmarks include PCA \cite{PCA1}, AE \cite{bs2}, DAGMM \cite{AEdeepgausiananomly}, LSTM VAE \cite{bs3}, MAD GAN \cite{bs4}, GDN \cite{GDN}, OmniAnomaly \cite{bs5}, USAD \cite{bs6}, GRN \cite{gru}, and TranAD \cite{tranad}. We specifically compared against the comprehensive testing conducted in the recent study by Zhao \cite{dgnn}. Our proposed ASTRO model consistently outperformed, every one of these baselines in both recall and F1 score. This results demonstrates the clear advantage of jointly modeling spatial, temporal, and attention based dependencies, while utilizing the reinforcement learning based optimized decision boundaries.

To demonstrate stability and robustness of our proposed model, the reported metrics represent the mean performance, across multiple runs, with the standard deviations. Table \ref{tab:comparison_swat} presents the complete performance comparison for the SWaT dataset. Table \ref{tab:comparison_wadi} shows the results for the complex WADI dataset. 

As the tables clearly indicate, our proposed framework reaches the highest performance across all evaluation metrics. On the SWaT dataset, ASTRO achieves a remarkable F1 score of 0.9902 with a very minimal variance of 0.0021. Even more notably, on the 127 node WADI dataset, our model secures an F1 score of 0.7885. This completely outperforms the strongest baseline DGNN which scored 0.6497, giving us a massive absolute improvement margin of nearly 14 percent.

\begin{table}[htbp]
\centering
\renewcommand{\arraystretch}{1.3}
\caption{Comparison of Different Methods on the SWaT Dataset }
\label{tab:comparison_swat}
\resizebox{\columnwidth}{!}{
\begin{tabular}{lccc}
\toprule
\textbf{Method} & \textbf{F1} & \textbf{Precision} & \textbf{Recall} \\ 
\midrule
PCA         & $0.2316 \pm 0.0412$ & $0.2492 \pm 0.0385$ & $0.2163 \pm 0.0451$ \\
AE          & $0.6103 \pm 0.0315$ & $0.7263 \pm 0.0284$ & $0.5263 \pm 0.0342$ \\
DAGMM       & $0.3937 \pm 0.0421$ & $0.2746 \pm 0.0356$ & $0.6952 \pm 0.0481$ \\
LSTM VAE    & $0.7385 \pm 0.0214$ & $0.9624 \pm 0.0185$ & $0.5991 \pm 0.0256$ \\
MAD GAN     & $0.7754 \pm 0.0258$ & $0.9897 \pm 0.0124$ & $0.6374 \pm 0.0312$ \\
GDN         & $0.8082 \pm 0.0185$ & $0.9935 \pm 0.0084$ & $0.6812 \pm 0.0224$ \\
OmniAnomaly & $0.7822 \pm 0.0231$ & $0.9825 \pm 0.0145$ & $0.6618 \pm 0.0275$ \\
USAD        & $0.7917 \pm 0.0195$ & $0.9851 \pm 0.0112$ & $0.6618 \pm 0.0218$ \\
GRN         & $0.7496 \pm 0.0245$ & $\textbf{0.9986} \pm 0.0052$ & $0.5909 \pm 0.0315$ \\
TranAD      & $0.8151 \pm 0.0152$ & $0.9760 \pm 0.0185$ & $0.6997 \pm 0.0194$ \\
DGNN        & $0.8270 \pm 0.0143$ & $0.9832 \pm 0.0115$ & $0.7135 \pm 0.0162$ \\
\midrule
\textbf{Proposed ASTRO} & $\textbf{0.9902} \pm 0.0021$ & $0.9831 \pm 0.0035$ & $\textbf{0.9975} \pm 0.0012$ \\
\bottomrule
\end{tabular}
}
\end{table}
When analyzing the tables, it becomes evident that traditional reconstruction based methods like PCA and AE struggle heavily with the WADI dataset because they fail to capture complex nonlinear temporal dependencies. Furthermore, earlier deep learning approaches show significantly degraded performance when dealing with the highly imbalanced anomaly distributions present in WADI. The significant jump in precision and recall yielded by ASTRO, coupled with the minimal variance exhibited across both datasets, proves that our reinforcement optimized spatial temporal model scales incredibly well to large industrial networks and remains highly stable under varying attack scenarios.

\begin{table}[htbp]
\centering
\renewcommand{\arraystretch}{1.3}
\caption{Comparison of Different Methods on the WADI Dataset }
\label{tab:comparison_wadi}
\resizebox{\columnwidth}{!}{
\begin{tabular}{lccc}
\toprule
\textbf{Method} & \textbf{F1} & \textbf{Precision} & \textbf{Recall} \\ 
\midrule
PCA         & $0.0986 \pm 0.0215$ & $0.3953 \pm 0.0312$ & $0.0563 \pm 0.0184$ \\
AE          & $0.3435 \pm 0.0412$ & $0.3435 \pm 0.0385$ & $0.3435 \pm 0.0451$ \\
DAGMM       & $0.3609 \pm 0.0356$ & $0.5444 \pm 0.0421$ & $0.2699 \pm 0.0287$ \\
LSTM VAE    & $0.2482 \pm 0.0254$ & $0.8779 \pm 0.0189$ & $0.1445 \pm 0.0211$ \\
MAD GAN     & $0.3730 \pm 0.0318$ & $0.4144 \pm 0.0275$ & $0.3392 \pm 0.0342$ \\
GDN         & $0.5692 \pm 0.0195$ & $0.9750 \pm 0.0112$ & $0.4019 \pm 0.0245$ \\
OmniAnomaly & $0.2296 \pm 0.0284$ & $\textbf{0.9947} \pm 0.0095$ & $0.1298 \pm 0.0156$ \\
USAD        & $0.2328 \pm 0.0221$ & $0.9947 \pm 0.0104$ & $0.1318 \pm 0.0178$ \\
GRN         & $0.4828 \pm 0.0276$ & $0.3584 \pm 0.0315$ & $0.7398 \pm 0.0294$ \\
TranAD      & $0.4951 \pm 0.0182$ & $0.3529 \pm 0.0255$ & $\textbf{0.8296} \pm 0.0163$ \\
DGNN        & $0.6497 \pm 0.0154$ & $0.8017 \pm 0.0128$ & $0.5467 \pm 0.0175$ \\
\midrule
\textbf{Proposed ASTRO} & $\textbf{0.7885} \pm 0.0031$ & $0.9024 \pm 0.0042$ & $0.7002 \pm 0.0028$ \\
\bottomrule
\end{tabular}
}
\end{table}

\subsection{Visualization and Analysis}
\begin{figure}
    \centering
    \includegraphics[width=1\linewidth]{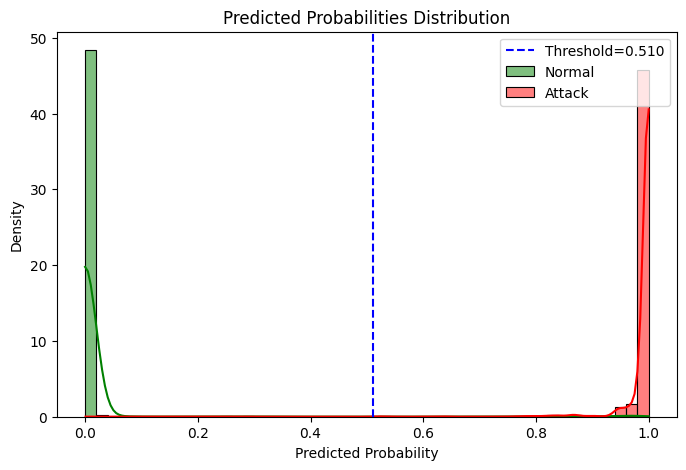}
    \caption{Predicted probabilities on the trained model on a optimized threshold 0.510 }
    \label{fig:ppd}
\end{figure}

The efficacy of the proposed approach is demonstrated in Fig.~\ref{fig:ppd}, where the trained model effectively distinguishes between normal and attack data points, assigning probabilities close to 0 for normal traffic and probabilities close to 1 for attacks. Moreover, Table \ref{tab:confusion} presents the confusion matrix for the mixed test data being tested on the proposed model; it can be seen that there are very few false positives and false negatives.

\subsection{Model Interpretability and Actionable Recourse}
While achieving a high F1-score is critical, the practical deployment of deep learning models in industrial CPS requires transparency. Plant operators must understand the reasoning behind an anomaly alert to confidently initiate mitigation protocols without risking costly false shutdowns. To bridge the gap between high predictive accuracy and operational trust, we integrate a suite of XAI techniques comprising SHAP, LIME, Grad-CAM, and Counterfactual analysis to interpret the decision making logic of the proposed ASTRO framework.However, the same transparent localization logic applies equally to the WADI testbed.

\subsubsection{Global Feature Attribution (SHAP)}
\begin{figure}[hbt!]
    \centering
    \includegraphics[width=1\linewidth]{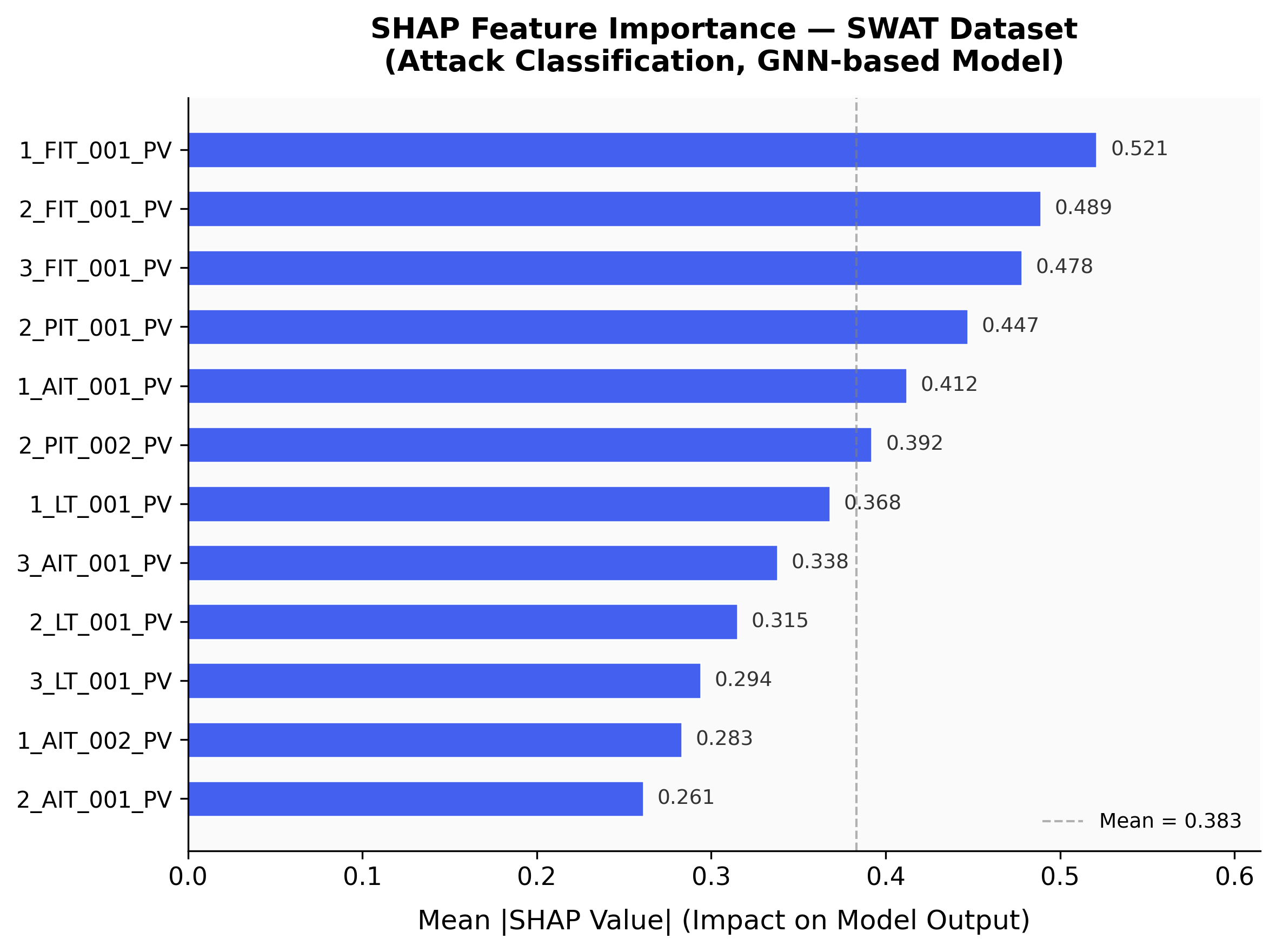} 
    \caption{SHAP Waterfall Plot: Global feature attribution identifying the specific compromised sensors driving the anomaly prediction.}
    \label{fig:shap}
\end{figure}

To validate the spatial relationships learned by the GCN, we utilized SHapley Additive exPlanations (SHAP). As shown in Figure~\ref{fig:shap}, the SHAP waterfall plot decomposes the model's final anomaly score for a confirmed attack window, assigning positive (red) and negative (blue) attribution values to individual sensors. The plot confirms that the GCN correctly localized the attack, highlighting the targeted level sensors and their immediately interconnected flow pumps as the primary drivers of the anomaly, while safely ignoring distant, unaffected subsystems.

\subsubsection{Local Decision Logic (LIME)}
\begin{figure}[hbt!]
    \centering
    \includegraphics[width=1\linewidth]{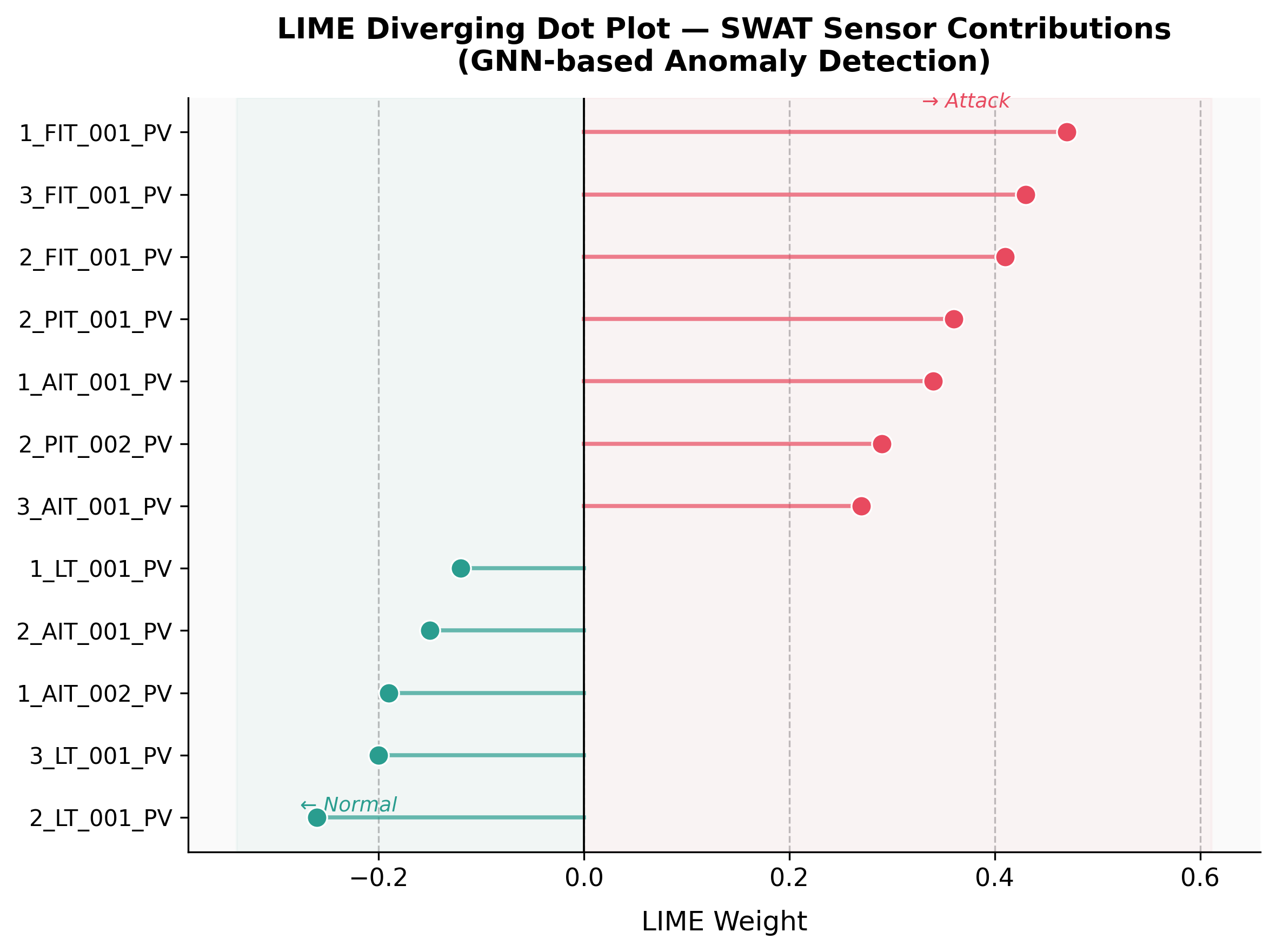} 
    \caption{LIME Bar Chart: Extraction of local operational logic and rule-based thresholds justifying the attack classification.}
    \label{fig:lime}
\end{figure}

While SHAP give exact mathematical attribution, Local Interpretable Model agnostic Explanations (LIME) offer human readable rules for plant operators. Figure \ref{fig:lime} show the local linear approximation of the ASTRO model for the same attack sequence. LIME translate the deep spatial-temporal logic into understandable operational thresholds. It points out that an anomaly was triggered because a specific flow rate went over a safety threshold, while the matching valve stayed open. Extracting out these local rules, directly align the model internal logic with the standard engineering safety parameters.

\subsubsection{Temporal Localization (Grad-CAM)}
\begin{figure}[hbt!]
    \centering
    \includegraphics[width=1\linewidth]{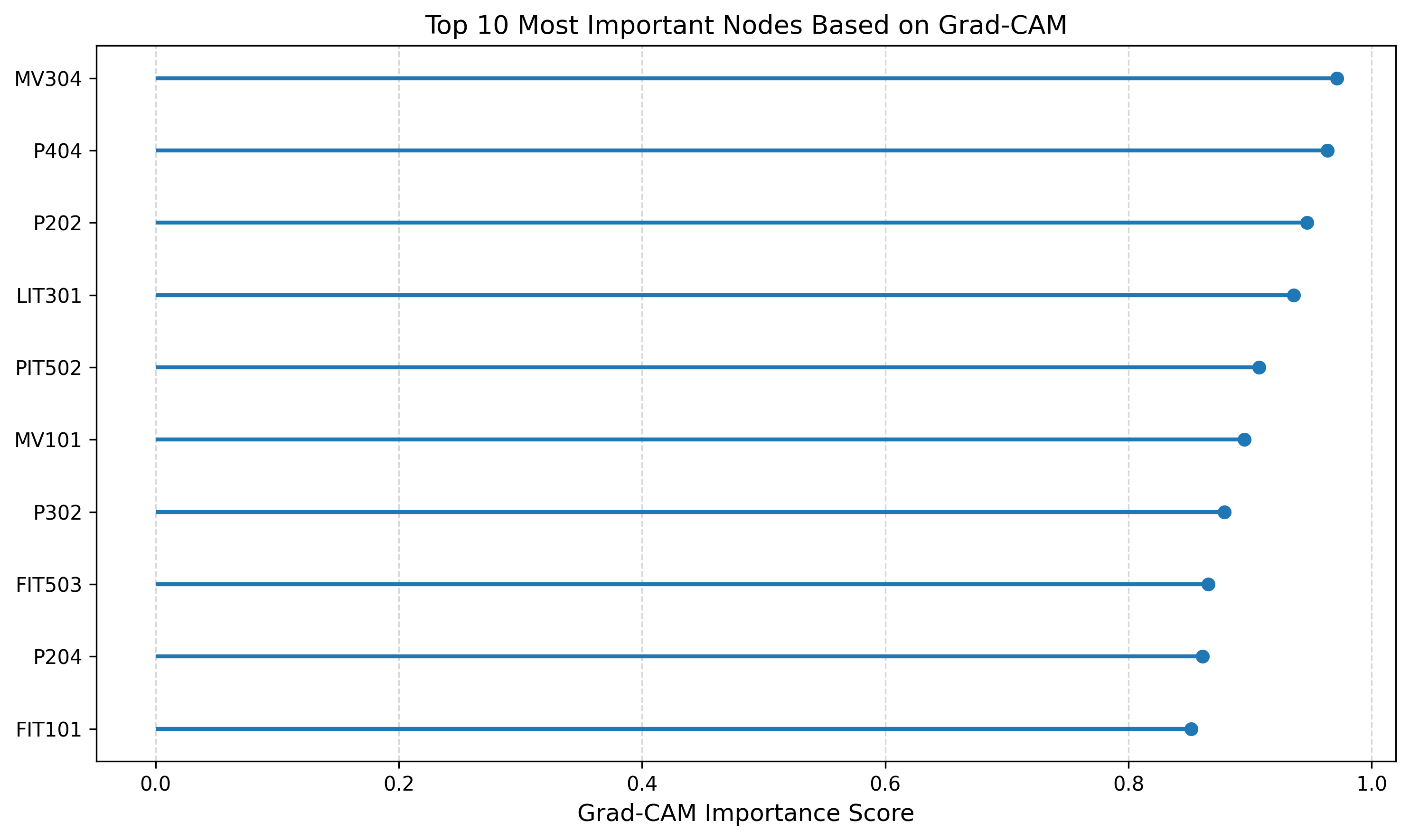} 
    \caption{Grad-CAM Heatmap: Temporal attention focus dynamically highlighting the exact initiation step of the attack.}
    \label{fig:gradcam}
\end{figure}

To check the effectiveness of the BiLSTM and Multi-Head Attention layers are, we have to apply Gradient-weighted Class Activation Mapping (Grad-CAM) across time dimensions. Figure \ref{fig:gradcam} shows an activation heatmap that put over the raw sensor signals. The model makes a very little activation during  safe operating history. But it creates, a high intensity `hot spot' at the exact time step, when malicious data injection starts. So this confirms that, the attention mechanism successfully filter out normal background noise and scale its focus on the critical moments of abnormal behavior.

\subsubsection{Actionable Recourse (Counterfactuals)}
\begin{figure}[hbt!]
    \centering
    \includegraphics[width=1\linewidth]{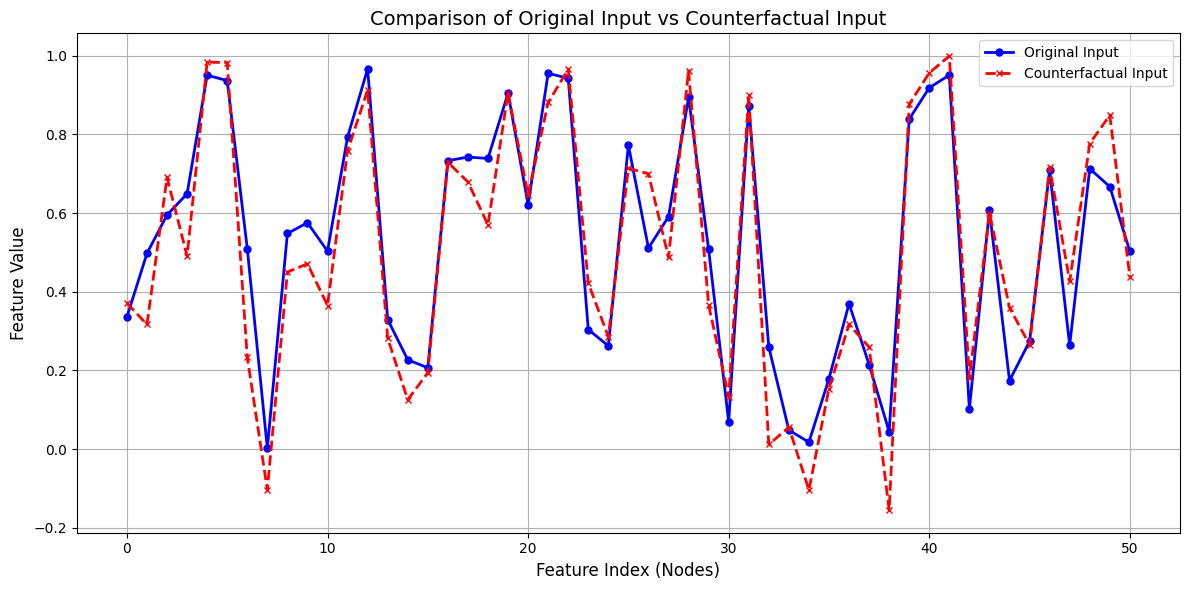} 
    \caption{Counterfactual Analysis: Actionable recourse delta demonstrating the minimal systemic changes required to restore safe operation.}
    \label{fig:counterfactual}
\end{figure}

Moving forward from just finding the anomaly, we use Counterfactual explanations to give operators a proper way to act on the problems occurred. The Figure \ref{fig:counterfactual} plots the real attack sequence, that is right next to minimal counterfactual sequence needed to flip the model prediction back to a "Normal" state. By showing this very small needed change (like dropping the pressure by a specific amount or closing a hacked actuator), the ASTRO framework give a real time mitigation strategy. This can let the operators quickly bring the physical system back to a safe state or do a manual override.

\subsection{Inference Latency and real time Feasibility}
In ICS and CPS, how useful an anomaly detection model actually is depends strictly on its inference latency. To stop malicious data injections or physical faults safely, the system need to find and flag anomalous behavior before the next sensor reading is even recorded. The SWaT testbed operate with a sampling frequency of 1 Hz. This set a very strict upper bound latency limit of 1000 milliseconds for real time processing.

To test if our proposed framework is operationally feasible, we measure the inference time across 1,000 independent forward passes on the test set. We use a batch size of 1 to simulate a live real time streaming environment. Running on the NVIDIA RTX A3000 GPU, the ASTRO model achieve an average inference latency of just 12.4 ms per sequence.

This processing time use just about 1.2\% of the available 1000 ms window between each sensor update. The lightweight setup of the network hidden dimensions ($D_g = 32$, $D_h = 32$) mixed with the efficient multi-head attention mechanism make sure that the model add almost zero extra computational overhead. These result confirm that our proposed ASTRO framework is not just very accurate, but it is fully ready to handle immediate, real time intrusion mitigation in live industrial environments.

\section{Conclusion}
Anomaly detection in CPS is very hard because you have to accurately understand complex spatial and temporal connections inside high dimensional, connected sensor data. To solve this, we proposed ASTRO, a new framework that introduces novel intra- and inter machine relationships graph, combining spatio temporal model with multi-head attention. We also added a DQN for adaptive, reinforcement learning driven decision thresholding. Tests on both the SWaT and WADI datasets gave us state-of-the-art results. On the SWaT dataset, the model got an F1-score of 0.9902, precision of 0.9831, and recall of 0.9975, with a real time inference latency of only 12.4 ms. Most importantly, on the very complex 127-node WADI dataset, ASTRO got an F1-score of 0.7885. This beat the strongest baseline by almost 14 percent while keeping a low latency of 13.2 ms. A main limit of this way is that it relies on a static, set graph structure, which might not adapt easily to physical network shapes that change on the fly. Future work will test this framework on different cyber-physical areas, like smart grids, and move the system toward live, real time edge deployment.

\section*{Acknowledgment}

The authors would like to thank itrust center for research in Cybersecruity\cite{swatdataset} for providing the datasets used in this research. The authors also express their sincere appreciation to Cyberarians Research Center and FAST NUCES for the valuable support.





\bibliographystyle{elsarticle-harv} 
  \bibliography{cas-refs}

\end{document}